\documentclass[11pt]{article}

\usepackage{graphicx}
\usepackage{amssymb}
\usepackage{amsmath}

\usepackage{lineno}
\usepackage{optidef}
\usepackage{adjustbox}
\usepackage{subcaption}
\usepackage{natbib}
\usepackage[superscript]{cite}
\usepackage{tabu}
\usepackage[flushleft]{threeparttable}
\usepackage{booktabs}
\usepackage{longtable}
\usepackage{multirow}
\usepackage{float}
\restylefloat{table}
\usepackage{geometry}
\usepackage{setspace}
\usepackage{adjustbox}
\usepackage{lscape}
\usepackage{bm}
\usepackage{xcolor}
\usepackage{authblk}

\date{ }
\begin{document}


\title{A Bayesian Finite Mixture Model with Variable Selection for Data with Mixed-type Variables}

\author[1,2]{Shu Wang}
\author[3,4,5]{Jonathan G. Yabes}
\author[3,4,5]{Chung-Chou H. Chang}
\affil[1]{\small{Department of Biostatistics, College of Public Health and Health Professions, University of Florida}}
\affil[2]{\small{University of Florida Health Cancer Center}}
\affil[3]{\small{Department of Biostatistics, Graduate School of Public Health, University of Pittsburgh}}
\affil[4]{\small{Department of Medicine, School of Medicine, University of Pittsburgh}}
\affil[5]{\small{Department of Clinical and Translational Science, School of Medicine, University of Pittsburgh}}

\renewcommand\Authands{ and }

\maketitle
\renewcommand{\abstractname}{\vspace{-\baselineskip}}

\label{firstpage}

\begin{abstract}

Finite mixture model is an important branch of clustering methods and can be applied on data sets with mixed types of variables. However, challenges exist in its applications. First, it typically relies on the EM algorithm which could be sensitive to the choice of initial values. Second, biomarkers subject to limits of detection (LOD) are common to encounter in clinical data, which brings censored variables into finite mixture model. Additionally, researchers are recently getting more interest in variable importance due to the increasing number of variables that become available for clustering.

To address these challenges, we propose a Bayesian finite mixture model to simultaneously conduct variable selection, account for biomarker LOD and obtain clustering results. We took a Bayesian approach to obtain parameter estimates and the cluster membership to bypass the limitation of the EM algorithm. To account for LOD, we added one more step in Gibbs sampling to iteratively fill in biomarker values below or above LODs. In addition, we put a spike-and-slab type of prior on each variable to obtain variable importance. Simulations across various scenarios were conducted to examine the performance of this method. Real data application on electronic health records was also conducted.

\end{abstract}



\section{Introduction}

The finite mixture model (FMM) \citep{mccutcheon1987latent, moustaki1996latent, nylund2007deciding} has been used to uncover the latent mixture probability distributions in a combined statistical distribution of a population \citep{deb2008finite} when the population is heterogeneous in characteristics and consisting with a combination of several more homogeneous subgroups. This method has a natural interpretation of heterogeneity through the mixture of finite components with a distributional assumption conditioning on components for each variable.

Unlike the commonly used nonparametric clustering algorithms (e.g., distance-based, density-based), the FMM technique is a model-based method that can easily handle variables with mixed variable types and can do variable selection while performing clustering. This is especially important when the number of variables involved is large.

The current method of FMM has to deal with several challenges. The first challenge is related to the use of the EM algorithm \citep{dempster1977maximum}, which is the mostly commonly used estimation procedure for the FMMs. In the EM algorithm the convergence rate could be very slow and the solution could be highly dependent on the choices of initial values, especially in the multivariate settings \citep{biernacki2003choosing,karlis2003choosing,mclachlan2007algorithm}.

The second challenge of using FMM is how to handle censored probability distributions. Oftentimes, covariates collected from medical setting include biomarker data of patients. Biomarkers are characteristics that can be accurately and reproducibly measured and accessed as an indicator of various biological processes \citep{biomarkers2001biomarkers,strimbu2010biomarkers}. Different biomarkers serve for different purposes. For example, temperature can be seen as a biomarker for fever; C-reactive protein (CRP) and Interleukin 6 (IL-6) are commonly used as biomarkers for sepsis \citep{pierrakos2010sepsis}. Therefore, biomarkers usually contain important diagnosis information about subjects. However, many biomarkers are subject to a limit of detection (LOD). LOD could be lower detection limit (i.e., values below this limit could not be measured), higher detection limit (i.e., values above this limit could not be measured), or both. When biomarkers are outcome variables in data analysis, biomarker values more extreme than the detection limit are usually represented by the detection limit value and an additional binary variable is included to indicating whether the corresponding value is actually undetected or measured. A semiparametric censored regression model can then be used \citep{tobin1958estimation,powell1984least,honore1992trimmed}. When biomarkers are predictors in analysis, multiple imputations are often used \citep{lubin2004epidemiologic,lee2012multiple,bernhardt2015statistical}. If the objective of the analysis is to cluster the feature space when data containing censored biomarkers, the two aforementioned approaches cannot be applied. Therefore, conducting clustering for data with LODs is still not well-addressed.

The objective of this paper is threefold. First, to overcome the estimation limitation of the EM algorithm, we adopt Gibbs sampling, which has been shown to be a valid and practical way in estimation of the FMMs \citep{geman1984stochastic}, \citep{diebolt1994estimation}. Second, we propose the use of a spike-and-slab type prior for categorical variables for the purpose of variable selection. Together with the traditional spike-and-slab prior for continuous variables, we incorporate variable selection into our Bayesian framework to provide quantitative information about variable importance. Third, we introduce an additional sampling step into our framework so that it is able to handle censored biomarker variables that are often encountered in clinical data.

Section~\ref{sec:review2} contains our review of currently used variable selection methods for the finite mixture models. We describe our proposed method in detail in Section~\ref{sec:method2}. Section~\ref{sec:sim2} includes simulation studies that are used to assess the performance of our methods and compare the performance to existing methods. In Section~\ref{sec:real2}, our proposed method is applied to identify sepsis phenotypes using demographic, clinical, and biomarker data collected from electronic health records. Our conclusions and the summary of this study are in Section~\ref{sec:dis2}.

\section{Existing methods review}
\label{sec:review2}

Overall, there are two categories of variable selection methods for supervised or unsupervised machine learning: filter methods and wrapper methods \citep{blum1997selection, guyon2003introduction, fop2018variable}.
Filter methods refer to those whose feature selection procedures are conducted separately with clustering procedures.
On the contrary, wrapper methods refer to those whose feature selection is conducted simultaneously with clustering procedures, like ``wrapped" around clustering procedures.
For example, the step 1 of the HyDaP algorithm \citep{2019arXiv190502257W} can be viewed as a filter method as it is conducted separately with the actual clustering algorithm. Wrapper methods are relatively more popular since it is naturally incorporated in clustering algorithms. In this section we focus on wrapper methods for FMMs.

Liu et al. proposed to conduct principle component analysis (PCA) before fitting Gaussian finite mixture models \citep{liu2003bayesian}. This method assumes that only the first $K$ factors are relevant to clustering, where $K$ is a random variable that has a prior distribution. However, factors having larger eigen values in PCA do not necessarily contain more important information for clustering \citep{chang1983using}.

Law et al. defined a binary indicator called feature saliency for each variable to reflect whether this variable is relevant to clustering or not. EM algorithm was used for estimation \citep{law2004simultaneous}. Let $\phi_m$ denote saliency for variable $m$. If $\phi_m=1$ then variable $m$ is relevant, otherwise variable $m$ is not, namely its distribution is independent of cluster labels.
Let $\bm{X}$ denote a data matrix with $n$ subjects and $M$ variables. Let $x_{im}$ denote variable $m$ of subject $i$. Let $G$ denote number of clusters. Let $\bm{Z}$ denote cluster indicator matrix; $z_{ig}$ denote indicator variable of subject $i$ belonging to cluster $g$. Let $\bm{\beta}$ denote distributional parameters, $\bm{\beta}_{mg}$ denote parameters of variable $m$ in cluster $g$, $\bm{\beta}_m$ denote marginal parameter of variable $m$. Then $L_m$, the likelihood due to variable $m$, is defined as:
\[
\begin{cases}
L_m|\bm{X},\bm{Z}, G, \bm{\beta}=\prod_{i=1}^{n}\prod_{g=1}^{G}[f(x_{im}|\bm{\beta}_{mg})]^{z_{ig}} & \quad \text{if variable } m \text{ is relevant}\\
L_m|\bm{X}, \bm{\beta}=\prod_{i=1}^{n}f(x_{im}|\bm{\beta}_m) & \quad \text{if variable } m \text{ is irrelevant}
\end{cases}
\]

Tadesse et al. and Li et al. used similar definitions of likelihood function and feature saliency. \citep{tadesse2005bayesian, li2009simultaneous}. Different with the one proposed by Law et al., the method proposed by Tadesse et al. detects discriminating variables through reversible-jump MCMC instead of EM algorithm for high-dimensional Gaussian finite mixture models.
Later, White et al. applied this idea on latent class analysis \citep{white2016bayesian} for Bayesian variable selection. While the method proposed by Li et al. adopted Variational Learning of Bayesian approximation (VB) for inference. They claimed that although using EM algorithm and VB usually lead to identical results, VB could avoid the situation of getting infinite likelihood when there is singular cluster, which may encounter using EM. Later Sun et.al. used the same framework and extended Gaussian mixture model to Student's t mixture model to better handle outliers \citep{sun2018simultaneous}.

Another category of methods is penalization approach \citep{fop2018variable}. The general idea is to maximize a penalized log-likelihood which is defined as:
$$l=\sum_{i=1}^{n}log\{\sum_{g=1}^{G}\tau_gf(\bm{x}_i|\bm{\beta}_g)\}-Q_\lambda(\bm{\beta})$$
where $\bm{x}_i$ is vector $(x_{i1},x_{i2},\dots,x_{iM})^T$; $\bm{\beta}$ is vector $(\bm{\beta}_1,\bm{\beta}_2,\dots,\bm{\beta}_G)^T$ and each $\bm{\beta}_g$ represents distributional parameters for cluster $g$; $Q$ is a function of distributional parameters; $\lambda$ is penalty parameter.
Different penalty functions were proposed including $L1$ penalty \citep{pan2007penalized}, sample size weighted $L1$ penalty \citep{bhattacharya2014lasso}, $L_\infty$ penalty \citep{wang2008variable} and other variations to achieve the goal of variable selection.
These methods all assume that the differences of mixture components lie in mean parameters, which assumes similar variance within all clusters.
This assumption was later relaxed by Xie et al. by adding two penalty terms so that the variance covariance structure is cluster-specific diagonal matrices \citep{xie2008penalized}.
Later this method was further extended to unconstrained variance covariance matrices \citep{zhou2009penalized}.

\begin{figure}[H]
     \centering
     \includegraphics[width=0.9\textwidth]{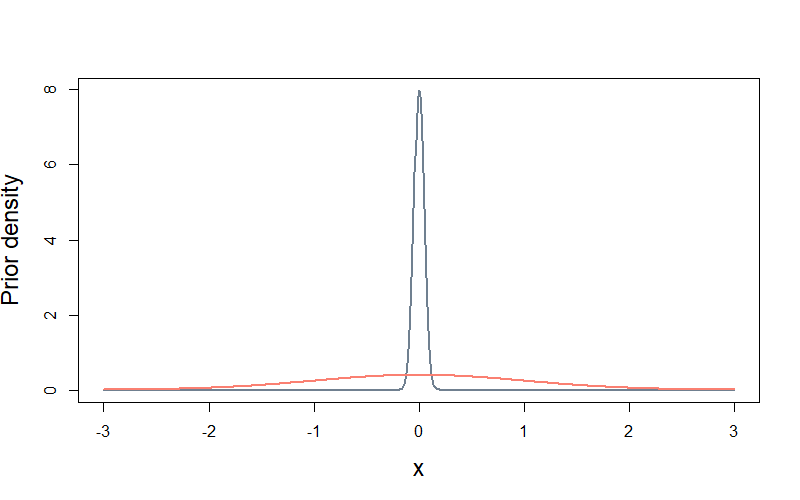}
     \caption{Demonstration of a spike and slab prior}
     \label{prior}
\end{figure}

Another category of methods that is commonly applied is spike and slab prior \citep{mitchell1988bayesian, madigan1994model, george1997approaches, ishwaran2005spike}. It is demonstrated in Figure~\ref{prior}.
Spike and slab prior is originally proposed for variable selection in linear regressions, but it can be naturally applied on Gaussian mixture models for unsupervised clustering.
This method first chooses one cluster as reference, and obtain mean difference between other clusters with this reference cluster. Similar to feature saliency, a binary indicator representing variable importance is also defined.
For other methods defined feature saliency, each variable has only one feature saliency indicating whether it is relevant to clustering or not.
While under spike and slab prior, each variable first has a cluster-specific importance, for cluster $2,3,\dots,G$ and then we aggregate these cluster-specific importance values into one overall importance or weight value.
Let $\Delta_{mg}$ denote importance of variable $m$ within cluster $g$, $\mu_{mg}$ the mean difference between cluster $g$ and cluster $1$ of variable $m$, where $g=2,3,\dots,G$.
$\Delta_{mg}=0$ indicates that for variable $m$, cluster $g$ is not different with reference cluster, namely variable $m$ within cluster $g$ is not important.
Then in next iteration, we assign $\mu_{mg}$ a ``spike" prior (the grey density in Figure~\ref{prior}): a Gaussian prior centered at 0 with very small variance. In this way, a value close to 0 is very likely to be sampled as the updated value of $\mu_{mg}$.
Otherwise if $\Delta_{mg}=1$, it indicates that for variable $m$, cluster $g$ is different with reference cluster, namely variable $m$ within cluster $g$ is important.
Then we assign $\mu_{mg}$ a ``slab" prior (the red density in Figure~\ref{prior}): a Gaussian prior centered at 0 with very large variance.
Therefore, any value is possible to be sampled as the updated value of $\mu_{mg}$. Let $\Delta_{m}$ denote final weight of variable $m$. Within each iteration, as long as $\Delta_{mg}=1$ for at least one cluster, we assign 1 to $\Delta_m$ for that iteration to represent that variable $m$ is important.
After all iterations, we calculate $p(\Delta_m=1)$ to be the weight of variable $m$.

Although researchers have explored a lot of variable selection methods for finite mixture models, all above methods only focus on single type of variables. Not many methods incorporate both types of variables. The method proposed by Raftery and Dean is able to handle various types of variables through stepwise regression like procedures \citep{raftery2006variable}. They used approximate Bayes factor as variable selection criteria and obtained final optimal variable sets through a greedy search algorithm.
This method naturally takes advantage of the parametric form of finite mixture models, but it could be extremely slow if a data set contains a large number of variables. Besides, this method only identifies that whether a variable is important or not instead of providing quantitative values for importance. Later it was extended through combining LASSO-like procedures and model selection procedures \citep{celeux2018variable} to be more efficient.
But similarly, this method does not provide real-valued variable weights.

\clearpage

\section{Bayesian finite mixture model with variable selection}
\label{sec:method2}

\subsection{Notation and proposed model}

We will first define some notations that will be used to define our proposed Bayesian finite mixture model.

Let $\bm{X}$, $\mathbf{x}_i=(x_{i1}, x_{i2},\dots, x_{iM})^T$ denote the vector of covariates for subject $i$ and $\bm{X}= (\mathbf{x}_1, \mathbf{x}_2, \dots, \mathbf{x}_n)^T$ be an $n \times M$ data matrix, where $n$ is the number of subjects and $M$ is the number of variables. Let $\mathbf{z}_i=(z_{i1}, z_{i2},\dots, z_{iG})^T$ denote the vector of cluster-membership indicators for subject $i$, where $z_{ig}=1$ if the subject $i$ belongs to cluster $g$ and $0$ otherwise. $\sum_{g=1}^{G}z_{ig}=1$, where $G$ is the number of clusters. Without loss of generality, we let the first $q$ variables to be continuous normally distributed and the rest $q+1^{th}$ to $M^{th}$ variables to be categorical.

Let $\bm{A}_1=(A_{11}, A_{12},\dots, A_{1q})^T $ denote the vector of mean values of all continuous variables in cluster $1$, where $A_{1m}$ is the mean of variable $m$ in cluster $1$, $m=1, 2,\dots,q$. We also define $\bm{\sigma}^2=(\sigma_1^2, \sigma_2^2,\dots, \sigma_q^2)^T$ as the vector of variances for all continuous variables, where $\sigma^2_{m}$ is the variance of variable $m$, $m=1, 2,\dots,q$.

We further define $\bm{\mu}$ as a $q \times G$ matrix with its $(m, g)$ element $\mu_{mg}$ representing the mean difference of variable $m$ between cluster $g$ and cluster $1$, where $m=1, 2,\dots,q$. Note that cluster $1$ is the reference cluster. Then, vector $\bm{\mu}_g=(\mu_{1g}, \mu_{2g},\dots, \mu_{qg})^T$ represents the mean difference between cluster $g$ and cluster $1$ of all normally distributed variables and vector $\bm{\mu}_m= (\mu_{m1}, \mu_{m2},\dots, \mu_{mG})^T$ represents the mean differences of variable $m$ for all $G$ clusters. For identifiability, $\bm{\mu}_1$ is set to be a vector with all elements being $1$.

For categorical variables, let $\bm{\theta}$ be a $(M-q) \times G$ matrix with its $(m, g)$ element $\bm{\theta}_{mg}$ representing the distributional parameters of variable $m$ within cluster $g$, where $m=q+1, q+2,\dots,M$. Vector $\bm{\theta}_g =(\bm{\theta}_{(q+1)g}, \bm{\theta}_{(q+2)g},\dots, \bm{\theta}_{Mg})^T$ represents the parameters of all categorical variables within cluster $g$ and vector $\bm{\theta}_m =(\bm{\theta}_{m1}, \bm{\theta}_{m2},\dots, \bm{\theta}_{mG})^T$ represents the parameters of variable $m$ for $m=q+1, q+2,\dots, M$.

We define density function of the finite mixture model (FMM) with $G$ clusters as

$$f(\mathbf {x}_i|\bm{\tau}, \bm{\theta}, \bm{A}_1,\bm{\mu}, \bm{\sigma}^2) =\sum_{g=1}^{G}\tau_gf(\mathbf{x}_i|\bm{\theta}_g,\bm{A}_1 ,\bm{\mu}_g,\bm{\sigma}^2),$$
where $\sum_{g=1}^{G}\tau_g=1$ and $\tau_g$ is the probability that a subject belongs to cluster $g$. We let $\bm{\tau}=(\tau_1, \tau_2,\dots, \tau_G)^T$ be the vector of cluster mixture probabilities.

For subject $i$, the corresponding density function for data $(\mathbf{x}_i$ given its cluster membership $\mathbf{z}_i)$ could be written as:
$$f(\mathbf{x}_i|\mathbf{z}_i, \bm{\theta},\bm{A}_1,\bm{\mu}, \bm{\sigma}^2)= \prod_{g=1}^{G}[f(\mathbf{x}_i|\bm{\theta}_g,\bm{A}_1,\bm{\mu}_g,\bm{\sigma}^2)]^{z_{ig}},$$
where $f(\mathbf{x}_i|\bm{\theta}_g,\bm{A}_1,\bm{\mu}_g,\bm{\sigma}^2)=\prod_{m=1}^{M}f(x_{im}|\theta_{mg}, A_{1m},\mu_{mg},\sigma_m^2)$ based on conditional independence, which assumes variables are independent with each other conditional on the cluster labels.

\subsection{Priors}

We define conjugate priors for all parameters specified in the above-defined FMM. Details of the prior distribution for each parameter is specified below.

For the cluster indicator matrix $\mathbf{Z}$, we specify a multinomial prior distribution with the form: $\mathbf{z}_i\sim Multinomial\left(G,\tau_1,\tau_2,\dots\tau_G\right)$. We also let the probability of cluster membership $\bm{\tau}$ follow a Dirichlet distribution
$$\bm{\tau}\sim Dir\left(\delta_1,\delta_2,\dots\delta_G\right),$$
where $\delta_1,\delta_2,\dots\delta_G$ are the hyper-parameters of $\bm{\tau}$.

Let parameter $A_{1m}$ follows $ N\left(\mu_A,\sigma_A^2\right)$, where $\mu_A$ and $\sigma_A^2$ are both hyper-parameters.

Let $\bm{\Delta}$ be a $G \times M$ matrix representing the collection of importance indicator for each variable, where $\Delta_{mg}$ is the importance indicator of variable $m$ within cluster $g$. Vector $\bm{\Delta}_g$ is $(\Delta_{1g}, \Delta_{2g},\dots, \Delta_{Mg})^T$ representing the importance indicator of all variables within cluster $g$; and vector $\bm{\Delta}_m$ is $(\Delta_{m1}, \Delta_{m2},\dots, \Delta_{mG})^T$ representing the importance indicator of variable $m$. For the purpose of variable selection, we apply a spike-and-slab priors for parameter $\mu_{mg}$ with the form
\[
\begin{cases}
\mu_{mg}\sim N\left(0, \sigma_{\Delta_0}^2\right) & \quad \text{if } \Delta_{mg}=0\\
\mu_{mg}\sim N\left(0, \sigma_{\Delta_1}^2\right) & \quad \text{if } \Delta_{mg}=1,
\end{cases}
\]
where $\sigma_{\Delta_0}^2$ and $\sigma_{\Delta_1}^2$ are both hyper-parameters. For identifiability, we define $\mu_{mg}=0$ when $g=1$. When $\Delta_{mg}=1$, namely cluster $g$ is different from cluster $1$, $\mu_{mg}$ should be away from 0. Therefore, $\sigma_{\Delta_1}^2$ should be a very large number, for increasing the variability of cluster mean difference, so that it is likely to be different with $0$. When $\Delta_{mg}=0$, which occurs if cluster $g$ is not different from cluster $1$, $\mu_{mg}$ should be close to 0. Therefore, $\sigma_{\Delta_0}^2$ should be a very small number so that the mean differences among cluster means would be close to zero. We used an Inverse-Gamma prior for parameter $\sigma_{\Delta_0}^2$:
$$\sigma_{\Delta_0}^2\sim Inv\Gamma\left(a_{\Delta_0},b_{\Delta_0}\right),$$
where $a_{\Delta_0}$ and $b_{\Delta_0}$ are both hyper-parameters.

The precision parameter $\gamma_{m}=\sigma_{m}^{-2}$ is assigned a prior following a Gamma distribution,
$$\gamma_{m}\sim \Gamma\left(\tilde{a},\tilde{b}\right),$$
where $\tilde{a}$ and $\tilde{b}$ are both hyper-parameters.

Inspired by the usual spike-and-slab priors for continuous variables, we propose similar priors for categorical variables. The challenge of building a spike-and-slab prior for a categorical variable include (1) we do not have an appropriate distribution for the probability difference; (2) a categorical variable may contain multiple levels so it is hard to compare all levels altogether between two clusters. Therefore, for a categorical variable, instead of comparing to a reference clustering group, we compare the distribution of the levels within each of the $G$ clusters to the overall marginal distribution of this categorical variable. If $\Delta_{mg}=0$, this indicates that the distribution of the categories within cluster $g$ is identical to the marginal distribution. If $\Delta_{mg}=1$, it implies that the distribution within cluster $g$ is different from the marginal distribution.

We define a Dirichlet distribution for categorical variable parameter $\bm{\theta}_{mg}$:
\[
\begin{cases}
\bm{\theta}_{mg}\sim Dir\left(\bm{\alpha}_{m\Delta_0}\right) & \quad \text{if } \Delta_{mg}=0\\
\bm{\theta}_{mg}\sim Dir\left(\bm{\alpha}_{\Delta_1}\right) & \quad \text{if } \Delta_{mg}=1
\end{cases}
\]
where $\bm{\alpha}_{m\Delta_0}$ and $\bm{\alpha}_{\Delta_1}$ are hyper-parameters.
Vector $\bm{\alpha}_{m\Delta_0}=(\alpha_{m\Delta_01},\alpha_{m\Delta_02},\dots,\alpha_{m\Delta_0L_m})^T$ is proportional to $\bm{\theta}_{mg}$ to make the corresponding prior center at marginal parameters of variable $m$, where $L_m$ is the number of categories of variable $m$; its elements are relatively larger numbers so that the prior is ``spike" at marginal parameters of variable $m$ when $\Delta_{mg}=0$.
Elements of vector $\bm{\alpha}_{\Delta_1}$ are all $1$ thus the pdf of this Dirichlet distribution is a constant.
Therefore, this prior becomes a ``slab" one when $\Delta_{mg}=1$.

We used Bernoulli distribution for the importance indicator variable $\Delta_{mg}$:
\[
\begin{cases}
\Delta_{mg}\sim Bern\left(p_{1m}\right) & \quad \text{if } m \in \{1,2,\dots,q\}\\
\Delta_{mg}\sim Bern\left(p_{2m}\right) & \quad \text{if } m \in \{q+1,q+2,\dots,M\}
\end{cases}
\]
where $p_{1m}$ and $p_{2m}$ are hyper-parameters of $\Delta_{mg}$.

Beta distribution for $p_{1m}$ and $p_{2m}$:
$$p_{1m}\sim Beta(a_{p_1},b_{p_1})$$
$$p_{2m}\sim Beta(a_{p_2},b_{p_2})$$
where $a_{p_1}$, $b_{p_1}$, $a_{p_2}$, $b_{p_2}$ are hyper-parameters.\\

\begin{figure}
\centering
\includegraphics[width=0.9\textwidth]{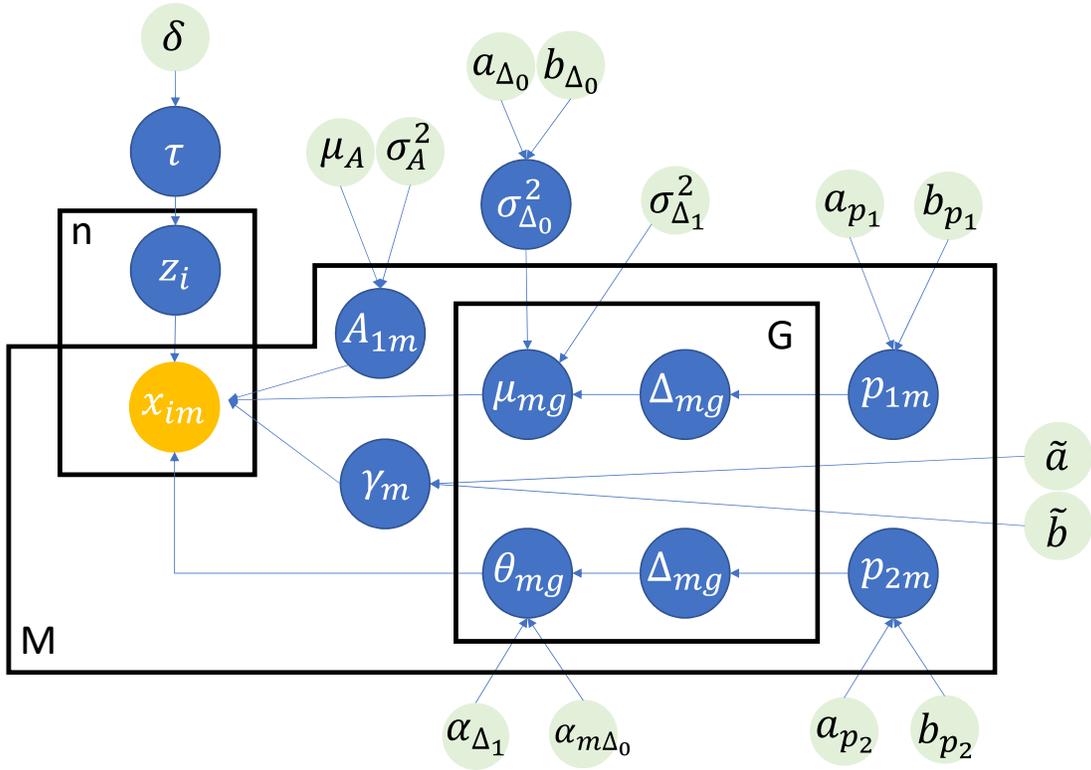}
\caption{Graphical representation of the proposed Bayesian FMM framework}
\label{bayes}
\end{figure}

A graphical depiction of our model is shown in Figure~\ref{bayes}.

\subsection{Posteriors}

With specified prior distributions of all parameters, the corresponding posterior distribution then can be obtained with the form:

\begin{align*}
f(Parameters|\mathbf{X}) &\propto \prod_{i=1}^{n}[\prod_{m=1}^{q}f(x_{im}|z_{im}, A_{1m}, \bm{\mu}_m,\gamma_{m})\prod_{m=q+1}^{M}f(x_{im}|z_{im},\bm{\theta}_m)]\\
& \quad \times\prod_{i=1}^{n}f(\mathbf{z}_i|\bm{\tau})f(\bm{\tau})\prod_{m=1}^{q}N(A_{1m};\mu_A, \sigma_A^2)\prod_{m=1}^{q}\Gamma(\gamma_m;\tilde{a},\tilde{b})\\
& \quad \times\prod_{m=1}^{q}\prod_{g=2}^{G}[N(\mu_{mg};0,\sigma_{\Delta_1}^2)\Delta_{mg}+N(\mu_{mg};0,\sigma_{\Delta_0}^2)(1-\Delta_{mg})]\\
& \quad \times Inv\Gamma(\sigma_{\Delta_0}^2;a_{\Delta_0},b_{\Delta_0})\prod_{m=1}^{q}\prod_{g=2}^{G}Bern(\Delta_{mg};p_{1m})\prod_{m=1}^{q}Beta(p_{1m};a_{p_1},b_{p_1})\\
& \quad \times\prod_{m=q+1}^{M}\prod_{g=1}^{G}[Dir(\theta_{mg};\bm{\alpha}_{\Delta_1})\Delta_{mg}+Dir(\theta_{mg};\bm{\alpha}_{m\Delta_0})(1-\Delta_{mg})]\\
& \quad \times\prod_{m=q+1}^{M}\prod_{g=1}^{G}Bern(\Delta_{mg};p_{2m})\prod_{m=q+1}^{M}Beta(p_{2m};a_{p_2},b_{p_2})
\end{align*}

We can derive posterior distribution for each parameter as below:

For cluster indicator matrix $\mathbf{Z}$:
\footnotesize{$$\bm{z}_i|\mathbf{x}_i,\bm{\tau},\bm{A_1},\bm{\mu},\bm{\gamma},\bm{\theta}\sim Multinomial\left(\frac{\tau_1f(\mathbf{x}_i|\bm{A_1},\bm{\mu}_1,\bm{\gamma}_1,\bm{\theta}_1)}{\sum_{g=1}^{G}\tau_gf(\mathbf{x}_i|\bm{A_1},\bm{\mu}_g,\bm{\gamma}_g,\bm{\theta}_g)}, \dots,\frac{\tau_Gf(\mathbf{x}_i|\bm{A_1},\bm{\mu}_G,\bm{\gamma}_G,\bm{\theta}_G)}{\sum_{g=1}^{G}\tau_gf(\mathbf{x}_i|\bm{A_1},\bm{\mu}_g,\bm{\gamma}_g,\bm{\theta}_g)}\right).$$}

\normalsize
For cluster mixture proportion $\bm{\tau}$:
$$\bm{\tau}|\mathbf{Z},\delta_1,\dots,\delta_G\sim Dir\left(\delta_1+\sum_{i=1}^{n}z_{i1},\dots, \delta_G+\sum_{i=1}^{n}z_{iG}\right).$$

For $A_{1m}$ in non-censored normal distributed variable $m$ within cluster $g$, where $m=1,2,\dots,q$:
\footnotesize{$$A_{1m}|\bm{X},\bm{Z},\bm{\mu}_m,\gamma_m,\mu_A,\sigma_A^2\sim N\left(\frac{\sigma_A^2(\sum_{i=1}^{n}x_{im}-\sum_{i=1}^{n}\sum_{g=1}^{G}z_{ig}\mu_{mg})+(\mu_A/\gamma_m)}{n\sigma_A^2+(1/\gamma_m)},\frac{\sigma_A^2/\gamma_m}{n\sigma_A^2+(1/\gamma_m)}\right).$$}

\normalsize
For mean difference $\mu_{mg}$, where $m=1,2,\dots,q$:\\
when $\Delta_{mg}=1$,
$$\mu_{mg}|\bm{X},\bm{Z},A_{1m},\gamma_m,\sigma_{\Delta_1}^2\sim N\left(\frac{\sigma_{\Delta_1}^2(\sum_{i=1}^{n}z_{ig}x_{im}-\sum_{i=1}^{n}z_{ig}A_{1m})}{\sigma_{\Delta_1}^2\sum_{i=1}^{n}z_{ig}+(1/\gamma_m)}, \frac{\sigma_{\Delta_1}^2/\gamma_m}{\sigma_{\Delta_1}^2\sum_{i=1}^{n}z_{ig}+(1/\gamma_m)}\right),$$
when $\Delta_{mg}=0$,
$$\mu_{mg}|\bm{X},\bm{Z},A_{1m},\gamma_m,\sigma_{\Delta_0}^2\sim N\left(\frac{\sigma_{\Delta_0}^2(\sum_{i=1}^{n}z_{ig}x_{im}-\sum_{i=1}^{n}z_{ig}A_{1m})}{\sigma_{\Delta_0}^2\sum_{i=1}^{n}z_{ig}+(1/\gamma_m)}, \frac{\sigma_{\Delta_0}^2/\gamma_m}{\sigma_{\Delta_0}^2\sum_{i=1}^{n}z_{ig}+(1/\gamma_m)}\right).$$

For precision parameter $\gamma_{m}$, where $m=1,2,\dots,q$:
$$\gamma_{m}|\mathbf{X},\mathbf{Z},A_{1m},\bm{\mu}_{m},\tilde{a},\tilde{b}\sim \Gamma\left(\tilde{a}+\frac{1}{2}n, \tilde{b}+\frac{1}{2}\sum_{i=1}^{n}\sum_{g=1}^{G}z_{ig}(x_{im}-A_{1m}-\mu_{mg})^2\right).$$

For hyper-parameter $\sigma_{\Delta_0}^2$ of $\mu_{mg}$:
$$\sigma_{\Delta_0}^2|\bm{\Delta},\bm{\mu},a_{\Delta_0},b_{\Delta_0}\sim Inv\Gamma\left(a_{\Delta_0}+\frac{1}{2}\sum_{m=1}^{q}\sum_{g=2}^{G}(1-\Delta_{mg}),b_{\Delta_0}+\frac{1}{2}\sum_{m=1}^{q}\sum_{g=2}^{G}(1-\Delta_{mg})\mu^2_{mg}\right).$$

For parameter $\bm{\theta}_{mg}$, where $m=q+1,q+2,\dots,M$:\\
When $\Delta_{mg}=1$,
$$\bm{\theta}_{mg}|\mathbf{X},\mathbf{Z},\bm{\alpha}_{\Delta_1}\sim Dir\left(\alpha_{\Delta_11}+\sum_{i=1}^{n}x_{im1}z_{ig},\dots,\alpha_{\Delta_1L_m}+\sum_{i=1}^{n}x_{imL_m}z_{ig}\right)$$
When $\Delta_{mg}=0$,
$$\bm{\theta}_{mg}|\mathbf{X},\mathbf{Z},\bm{\alpha}_{m\Delta_0}\sim Dir\left(\alpha_{m\Delta_01}+\sum_{i=1}^{n}x_{im1}z_{ig},\dots,\alpha_{m\Delta_0L_m}+\sum_{i=1}^{n}x_{imL_m}z_{ig}\right).$$

For importance indicator variable $\bm{\Delta}$:\\
When $m \in {1,2,\dots,q}$:
$$\Delta_{mg}|\mu_{mg},p_{1m},\sigma_{\Delta_0}^2,\sigma_{\Delta_1}^2\sim Bernoulli \left(\frac{p_{1m}N(\mu_{mg};0,\sigma_{\Delta_1}^2)}{p_{1m}N(\mu_{mg};0,\sigma_{\Delta_1}^2)+(1-p_{1m})N(\mu_{mg};0,\sigma_{\Delta_0}^2)}\right)$$
When $m \in {q+1,q+2,\dots,M}$:
$$\Delta_{mg}|\mu_{mg},p_{2m},\bm{\alpha}_{m\Delta_0},\bm{\alpha}_{\Delta_1}\sim Bernoulli \left(\frac{p_{2m}Dir(\theta_{mg};\bm{\alpha}_{\Delta_1})}{p_{2m}Dir(\theta_{mg};\bm{\alpha}_{\Delta_1})+(1-p_{2m})Dir(\theta_{mg};\bm{\alpha}_{m\Delta_0})}\right).$$

For $p_{1m}$ for continuous variables, where $m=1,2,\dots,q$:
$$p_{1m}|\bm{\Delta}_m,a_{p_1},b_{p_1} \sim Beta \left(a_{p_1}+\sum_{g=2}^{G}\Delta_{mg}, b_{p_1}+\sum_{g=2}^{G}(1-\Delta_{mg})\right).$$

For $p_{2m}$ for categorical variables, where $m=q+1,q+2,\dots,M$:
$$p_{2m}|\bm{\Delta}_m,a_{p_2},b_{p_2} \sim Beta \left(a_{p_2}+\sum_{g=1}^{G}\Delta_{mg}, b_{p_2}+\sum_{g=1}^{G}(1-\Delta_{mg})\right).$$
We set different hyper-parameters for $p_{1m}$ and $p_{2m}$ to make variable selection more flexible. In this way, we can control the extent of shrinkage for continuous and categorical variables separately. In the next section, we will introduce how different choices of hyper-parameters could affect variable selection with more details.

For continuous variables with detection limit, we add one more sampling step: sample values below the lower detection limit and above the upper detection limit from the remainder of truncated normal distributions, respectively.
$$f(x_{i^{\prime}m}|A_{1m},\mu_{mg}, \gamma_{m})=\left(\frac{\sqrt{\gamma_{m}}h(x_{i^{\prime}m})}{\Phi[\sqrt{\gamma_{m}}(C_{Lm}-\mu_{mg}-A_{1m})]}\right)^{z_{ig}}$$
$$f(x_{i^*m}|A_{1m},\mu_{mg}, \gamma_{m})=\left(\frac{\sqrt{\gamma_{m}}h(x_{i^*m})}{1-\Phi[\sqrt{\gamma_{m}}(C_{Um}-\mu_{mg}-A_{1m})]}\right)^{z_{ig}},$$
where $C_{Um}$ is the upper limit and $C_{Lm}$ is the lower limit for variable $m$; $i^{\prime}$ represents subjects whose real values of variable $m$ are lower than $C_{Lm}$; $i^*$ represents subjects whose real values of variable $m$ are higher than $C_{Um}$; $h$ is standard normal distribution; $\Phi$ is CDF of standard normal distribution.

\subsection{Hyper-parameters}

From the Bayesian framework of our proposed model, we will need to specify the values of the following hyper-parameters before starting the sampling procedures: $\bm{\delta}$ for $\bm{\tau}$; $\mu_A$ and $\sigma_A^2$ for $\bm{A_1}$; $a_{\Delta_0}$ and $b_{\Delta_0}$ for $\sigma_{\Delta_0}^2$; $\sigma_{\Delta_1}^2$ for $\bm{\mu}$; $\tilde{a}$ and $\tilde{b}$ for $\bm{\gamma}$; $\bm{\alpha}_{\Delta_1}$ and $\bm{\alpha}_{m\Delta_0}$ for $\bm{\theta}$; $a_{p_1}$ and $b_{p_1}$ for $p_{1m}$; and $a_{p_2}$ and $b_{p_2}$ for $p_{2m}$.

\begin{figure}[H]
\centering
\includegraphics[width=0.9\textwidth]{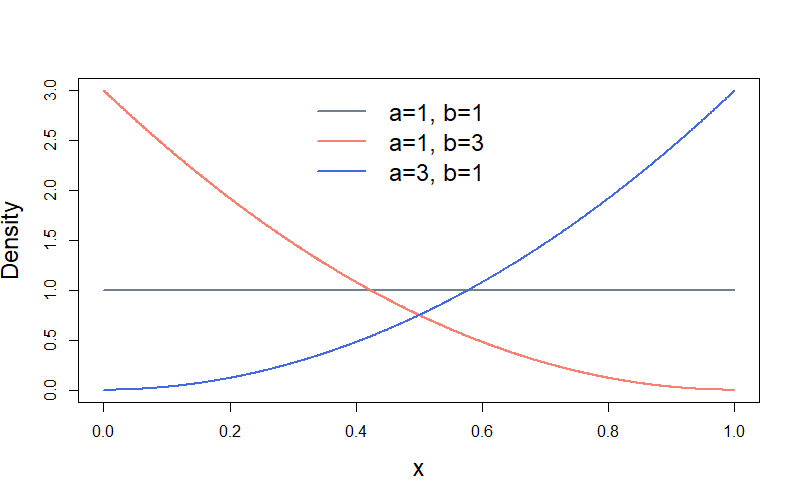}
\caption{Beta priors}
\label{beta}
\end{figure}

We can control the extent of shrinkage through controlling hyper-parameters $a_{p_1}$ and $b_{p_1}$ for continuous variables, and $a_{p_2}$ and $b_{p_2}$ for categorical variables.
We can start with $a_{p_1}=b_{p_1}=a_{p_2}=b_{p_2}=1$, namely $Beta(1,1)$ to be the prior distribution of $p_{1m}$ and $p_{2m}$.
If we prefer more shrinkage, then we could increase $b_{p_1}$ and $b_{p_2}$; otherwise we could decrease $b_{p_1}$ and $b_{p_2}$.
In Figure~\ref{beta} we show probability density functions of $Beta(1,1)$, $Beta(1,3)$ and $Beta(3,1)$ as an example.
We can find that $Beta(1,1)$ (the grey line in Figure~\ref{beta}) is the same as $Unif(0,1)$, namely it is a non-informative flat prior.
Under prior $Beta(1,3)$ (the red curve in Figure~\ref{beta}), it is more likely to sample smaller numbers for corresponding $\Delta_{mg}$ so that we can achieve more shrinkage.
While under prior $Beta(3,1)$ (the blue curve in Figure~\ref{beta}), it is more likely to sample larger numbers for the corresponding $\Delta_{mg}$ so that we put less shrinkage.
Assigning different hyper-parameters to $p_{1m}$ and $p_{2m}$ makes the shrinkage more flexible since we can control continuous and categorical variable separately.
If we prefer the same shrinkage for all variables, we can simply set $a_{p_1}=a_{p_2}$ and $b_{p_1}=b_{p_2}$.


\subsection{Estimation procedure}

Since all parameters in the model have conjugate priors, we can apply Gibbs sampling to estimate unknown parameters and the procedure is summarized as follows.

Step 1: Set initial values for all parameters.

Step 2: For each variable in the dataset (e.g., age), update its distributional parameters (e.g., mean and standard deviation), namely $\bm{A_1}$, $\bm{\mu}$, and $\bm{\gamma}$ for a continuous variable or $\bm{\theta}$ for a categorical variable by sampling from the corresponding posterior distributions of parameters. If censored biomarker variables are encountered, update the censored variables before updating the distributional parameters.

Step 3: For each variable, update $\bm{\Delta}$ given the values of all other parameters.

Step 4: For each variable, update $p_{1m}$ or $p_{2m}$ given the values of all other parameters.

Step 5: Update $\sigma_{\Delta_0}^2$, $\mathbf{Z}$, and $\bm{\tau}$ given the values of all other parameters.

Step 6: Repeat Steps 2 to 5 for many iterations.

Step 7: Estimate each parameter using the mean value of its posterior distribution. Also, calculate the posterior probabilities of cluster membership for each subject and assign the cluster with the highest probability to that subject.

\subsection{Label switching in Gibbs sampling}

Label switching is a common problem in Gibbs sampling. It has been extensively studied and discussed in the literatures \citep{diebolt1994estimation,stephens2000dealing,fruhwirth2001markov,marin2005bayesian,papastamoulis2010artificial}.
 Scrambling the cluster labels will not affect the likelihood function, the priors, and the posterior distributions. In Gibbs sampling of the cluster-specific parameters, there is a possibility of assigning wrong cluster labels to these parameters. This is called the label switching issue, which could cause biased estimation of the cluster-specific parameters. We adopt Stephen's method \citep{stephens2000dealing,papastamoulis2015label} to resolve this issue.

Let $\bm{P}^{(t)}$ be a $n\times G$ matrix with element $p^{(t)}_{ig}$ representing posterior probabilities of a subject $i$ belonging to cluster $g$ at MCMC iteration $t$; $t=1,2,\dots,T$, where $n$ is total number of subjects, $T$ is the total MCMC iteration times and $G$ is the total number of clusters. The basic idea of Stephen's method is trying to permute the estimates from the MCMC iterations if needed so that those posterior probability matrix $\bm{P}^{(t)}$ agree each other for $t=1,2,\dots,T$. Specifically, Stephen's Method minimizes the Kullback-Leibler divergence (also called \textit{relative entropy}) between $\frac{1}{T}\sum_{t=1}^{T}\bm{P}^{(t)}$ and $\bm{P}^{(t)}$ for each $t$.\\~\\
The relabeling algorithm is summarized as follows:

1. Assign an initial cluster labeling for all the MCMC iterations. We let $\bm{U}$ be a $T \times G$ matrix whose $t$th row represents the cluster labels for the $t$th iteration. For example, if $t=1$ and $G=3$, the cluster labels for the first MCMC iteration could be $(1, 2, 3)$, $(1, 3, 2)$, $(2, 1, 3)$, $(2, 3, 1)$, $(3, 1,2)$ or $(3, 2, 1)$. To assign an initial clustering labeling is equivalent to assigning an initial values of the matrix $\bm{U}$.

2. Based on the current cluster labels, we calculate the mean posterior probability of cluster membership for all subjects across all iterations by averaging $\bm{P}^{(1)}, \bm{P}^{(2)}, \dots, \bm{P}^{(T)}$. That is, the mean posterior probability of the $g$th cluster membership for subject $i$ has the form:
$$\bar{p}_{ig}=\frac{1}{T}\sum_{t=1}^{T}p^{(t)}_{ig},$$ for $i=1,2,...,n$ and $g=1,2,...,G,$
where $p_{ig}^{(t)}$ is the $(i,g)$th element of the matrix $\bm{P}^{(t)}$.

3. Update each row of the matrix $\bm{U}$, the cluster labels of an iteration, by minimizing
$$\sum_{i=1}^{n}\sum_{g=1}^{G}p^{(t)}_{ig}\log(\frac{p^{(t)}_{ig}}{\bar{p}_{ig}}),$$for $t=1,2,\dots,T$.

4. Repeat Steps 2 and 3 until matrix $\bm{U}$ is unchanged.
Let $\bm{\Psi}$ be the $T \times G$ matrix of a parameter that needs to be estimated, where $T$ is the total number of MCMC iterations and $G$ is the total number of clusters. We will obtain the final parameter estimates for each iteration (each row of $\bm{\Psi}$) by permuting the cluster labels based on the values in the matrix $\bm{U}$.

\section{Simulation studies}
\label{sec:sim2}

In this section we use simulations to evaluate the performance of the proposed Bayesian FMM relative to the other existing approaches. 
As reviewed in Section~\ref{sec:review2}, most existing FMM frameworks with the ability of variable selection can only be applied on data consists of single variable type. Therefore, we chose methods that can handle mixed data: the HyDaP algorithm \citep{2019arXiv190502257W}, Partition Around Medoids (PAM) with Gower distance \citep{gower1971general}, K-prototypes \citep{huang1998extensions}, regular FMM that uses the EM algorithm for estimation, and PAM with distance defined in factorial analysis of mixed data (FAMD) \citep{pages2014multiple} as comparison methods. Clustering performance is evaluated by Adjusted Rand Index (ARI).

Assuming that there are 3 underlying true clusters with cluster sizes of 100, 100, and 100. In terms of variable importance, we designed 4 scenarios covering (1) both variable types contribute to clustering, (2) only continuous variables contribute to clustering, and (3) only categorical variables contribute to clustering. They are similar to the settings used in the paper that proposed the HyDaP algorithm. Details of these simulation settings can be found in Table~\ref{tab:setting}.
To evaluate clustering performance of different methods in the presence of censored biomarkers, we focused on the first simulation scenario and examined combinations of varying censoring proportions and varying numbers of censored variables.
For each simulation scenario, 500 datasets were generated.

\begin{table}
\begin{adjustbox}{center}
\footnotesize
\centering
\resizebox{0.95\textwidth}{!}{%
\begin{tabular}{c c c c c c c}
\toprule
 \footnotesize{\textbf{Variable}} & \footnotesize{\textbf{Cluster$^a$}} & \footnotesize{\textbf{Sim 1(a)}} & \footnotesize{\textbf{Sim 1(b)}} & \footnotesize{\textbf{Sim 2}} & \footnotesize{\textbf{Sim 3}}\\
\midrule
  & 1 & $\bm{N(-2, 2)}^b$ &$\bm{N(-2, 2)}$ & $\bm{N(-2, 2)}$ & \\
 $x_1$ & 2 & $\bm{N(2, 2)}$ & $\bm{N(-1, 2)}$ &$\bm{N(2, 2)}$ & $N(0, 0.5)$\\
 & 3 & $\bm{N(6, 2)}$ & $\bm{N(0, 2)}$&$\bm{N(6, 2)}$ & \\
\midrule
  & 1 & $\bm{N(20, 1)}$ &$\bm{N(20, 1)}$ &$\bm{N(20, 1)}$ & \\
 $x_2$ & 2 & $\bm{N(25, 1)}$ &$\bm{N(24, 1)}$ &$\bm{N(25, 1)}$ &$N(-3, 1)$\\
 & 3 & $\bm{N(18, 1)}$ &$\bm{N(21, 1)}$ &$\bm{N(18, 1)}$ & \\
\midrule
 & 1 & $\bm{N(0, 1)}$ &$\bm{N(5, 1)}$ & $\bm{N(0, 1)}$ & \\
 $x_3$ & 2 & $\bm{N(-7, 1)}$ &$\bm{N(8, 1)}$ &$\bm{N(-7, 1)}$ &$N(4, 2)$\\
 & 3 & $\bm{N(4, 1)}$ &$\bm{N(7, 1)}$ &$\bm{N(4, 1)}$ & \\
 \midrule
  & 1 & & & & \\
 $x_4$ & 2 & $N(0, 1)$ & $N(0, 1)$&$N(0, 1)$ &$N(0, 1)$\\
 & 3 & & & & \\
 \midrule
 & 1 & $\bm{M(0.1, 0.1, 0.8)}$ & $\bm{N(-1, 1)}$ &$M(0.3, 0.3, 0.4)$ & $\bm{M(0.05, 0.05, 0.9)}$ \\
 $x_5$ & 2 & $\bm{M(0.1, 0.8, 0.1)}$ &$\bm{N(1, 1)}$ & $M(0.3, 0.3, 0.4)$ & $\bm{M(0.05, 0.9, 0.05)}$\\
 & 3 & $\bm{M(0.8, 0.1, 0.1)}$ & $\bm{N(-2, 1)}$ &$M(0.4, 0.3, 0.3)$ & $\bm{M(0.9, 0.05, 0.05)}$\\
 \midrule
 & 1 & &$\bm{N(0, 1)}$ & & $M(0.3, 0.3, 0.4)$\\
 $x_6$ & 2 & & $\bm{N(-1, 1)}$ &  &$M(0.4, 0.3, 0.3)$\\
 & 3 & &$\bm{N(2, 1)}$ & & $M(0.3, 0.4, 0.3)$\\
 \midrule
 & 1 & &$\bm{N(2, 1)}$ & & $\bm{M(0.9, 0.05, 0.05)}$\\
 $x_7$ & 2 & &$\bm{N(1, 1)}$ & &$\bm{M(0.05, 0.9, 0.05)}$\\
 & 3 & &$\bm{N(0, 1)}$ & & $\bm{M(0.05, 0.05, 0.9)}$\\
 \midrule
 & 1 & &$\bm{M(0.05, 0.05, 0.9)}$ & & & \\
 $x_8$ & 2 & &$\bm{M(0.05, 0.9, 0.05)}$ & & &\\
 & 3 & &$\bm{M(0.9, 0.05, 0.05)}$ & & & \\ 
 \midrule
 & 1 & &$M(0.3, 0.3, 0.4)$ & & & \\
 $x_9$ & 2 & &$M(0.4, 0.3, 0.3)$ & & &\\
 & 3 & &$M(0.3, 0.4, 0.3)$ & & & \\
 \midrule
 & 1 & &$\bm{M(0.9, 0.05, 0.05)}$ & & & \\
 $x_{10}$ & 2 & &$\bm{M(0.05, 0.9, 0.05)}$ & & &\\
 & 3 & &$\bm{M(0.05, 0.05, 0.9)}$ & & & \\ \midrule
  \multicolumn{6}{l}{$^a$Sample sizes for 3 clusters are 100, 100 and 100; $^b$variables with bolded distributions are important in clustering} \\ \bottomrule
 \end{tabular}%
 }
\caption{Simulation settings}
\label{tab:setting}
\end{adjustbox}
\end{table}

Our hyper-parameters used in simulations are as follows: $(0.33,0.33,0.33)^T$ was used for $\bm{\delta}$ assuming we do not have much informative on $\bm{\tau}$; 0 and 100 were used for $\mu_A$ and $\sigma_A^2$ assuming we do not have much informative on $A_{1m}$; 2 and 0.0001 were used for $a_{\Delta_0}$ and $b_{\Delta_0}$ as we believe $\sigma_{\Delta_0}^2$ should be a small number so that the corresponding prior for $\mu_{mg}$ is a ``spike" one; 1000 was used for $\sigma_{\Delta_1}^2$ as we believe $\sigma_{\Delta_1}^2$ should be a large number so that the corresponding prior for $\mu_{mg}$ is a ``slab" one; 2 and 1 were used for $\tilde{a}$ and $\tilde{b}$ assuming we do not have much informative on $\gamma_m$; 1 for $\bm{\alpha}_{\Delta_1}$ so that the corresponding prior for $\theta_{mg}$ is a ``slab" one.
We used $(10, 10, 10)^T$ for $\bm{\alpha}_{m\Delta_0}$, where $m=1,2,\dots,M$, since every categorical variable has 3 categories with true marginal probabilities $(0.33, 0.33, 0.33)^T$.
We used 1 and 2 for $a_{p_1}$ and $b_{p_1}$, 1 and 2 for $a_{p_2}$ and $b_{p_2}$ as we would like some extent of shrinkage on both continuous and categorical variables.

\subsection{Without existence of censored biomarker variables}
Clustering performances in terms of ARI are shown in Table~\ref{tab:perform2}. The first row, Bayesian FMM, is our proposed Bayesian FMM with variable selection. In addition, the median and $2.5^{th}$ to $97.5^{th}$ percentile interval of variable weights obtained from our proposed Bayesian FMM with variable selection across all simulated datasets are shown in Table~\ref{tab:weight}.

In simulation 1(a), we simulated a total of 5 variables: 4 continuous and 1 categorical. All except one continuous variable truly contribute to clustering. The sole categorical variable also contributes to clustering.
Our Bayesian FMM with variable selection performed the best (ARI 1.00 [0.98, 1.00]) compared with other methods.
Table~\ref{tab:weight} shows that the median weights of variables $x_1$, $x_2$, $x_3$ and $x_5$ are $1.00$ with $2.5^{th}$ and $97.5^{th}$ percentile interval $(1.00, 1.00)$, indicating that these variables were found to be important for clustering. Meanwhile, variable $x_4$ had low weights (0.01 [0.01, 0.06]). These weights correspond to the true setting.

In simulation 1(b), we simulated a total of 10 variables: 7 continuous and 3 categorical. Six out of seven continuous variables truly contribute to clustering; two out of three categorical variables contribute to clustering. The difference between this setting and 1(a) is that in simulation 1(a), continuous variables with dominant influences exist while such variables do not exist in simulation 1(b).
Our Bayesian FMM with variable selection performed the best (ARI 0.99 [0.96, 1.00]) compared with other methods.
In Table~\ref{tab:weight}, variables $x_1$, $x_4$ and $x_7$ had weights 0.05 (0.04, 0.07), 0.04 (0.03, 0.04), 0.05 (0.04, 0.07) indicating none of them has contribution to clustering; variables $x_3$, $x_5$, $x_6$ and $x_9$ had weights 0.10 (0.05, 0.19), 0.12 (0.06, 0.16), 0.11 (0.06, 0.17) and 0.12 (0.08, 0.22) respectively indicating that they have small contributions to clustering; variable $x_2$ had weight 0.30 (0.11, 0.41) indicating that it had slightly larger importance than $x_3$, $x_5$, $x_6$ and $x_9$.
Variables $x_8$ and $x_{10}$ had weights 1.00 (1.00, 1.00) indicating they're dominant variables that are highly relevant to clustering. These weights correctly reflect the true setting.

\begingroup
\renewcommand{\arraystretch}{1.4}
\begin{table}[H]
\footnotesize
\begin{adjustbox}{center}
\centering
\begin{tabular}{@{}ccccc@{}}
\toprule
\textbf{Clustering Method} & \multicolumn{4}{c}{\textbf{ARI, median (2.5th percentile, 97.5th percentile)}} \\ \midrule
 & Sim 1(a) & Sim 1(b) & Sim 2(a) & Sim 3 \\ \cmidrule(l){2-5}
Bayesian FMM & 1.00 (0.98, 1.00) & 0.99 (0.96, 1.00) & 1.00 (0.98, 1.00) & 0.74 (0.65, 0.83) \\
HyDaP & 0.97 (0.92, 1.00) & 0.95 (0.87, 1.00) & 0.98 (0.94, 1.00) & 0.75 (0.63, 0.85) \\
PAM + Gower distance & 0.70 (0.58, 0.80) & 0.87 (0.76, 0.96) & 0.01 (-0.01, 0.04) & 0.71 (0.31, 0.84) \\
K-prototypes & 1.00 (0.96, 1.00) & 0.93 (0.79, 1.00) & 0.98 (0.92, 1.00) & 0.17 (-0.01, 0.26) \\
Finite mixture model & 1.00 (0.98, 1.00) & 0.98 (0.44, 1.00) & 1.00 (0.56, 1.00) & 0.72 (0.56, 0.85) \\
PAM + FAMD distance & 0.78 (0.66, 0.89) & 0.93 (0.84, 0.98) & 0.09 (-0.01, 0.44) & 0.73 (0.22, 0.84) \\ \bottomrule
\end{tabular}
\caption{Performance comparison in different simulation settings}
\label{tab:perform2}
\end{adjustbox}
\end{table}
\endgroup

In simulation 2, we simulated a total of 5 variables: 4 continuous and 1 categorical. This setting is the same as setting 1(a) except that the sole categorical variable does not contribute to clustering.
Our Bayesian FMM with variable selection performed the best (ARI 1.00 [0.98, 1.00]) compared with other methods.
Variables $x_1$ to $x_4$ exhibited similar weights as those in simulation 1(a), but variable $x_5$ now had a low weight (0.12 [0.08, 0.23]), reflecting the underlying setting.

In simulation 3, we simulated a total of 7 variables: 4 continuous and 3 categorical. None of the continuous variables truly contribute to clustering. Two out of three categorical variables contribute to clustering.
Our Bayesian FMM with variable selection performed similarly (ARI 0.74 [0.65, 0.83]) with the best performer: the HyDaP algorithm (ARI 0.75 [0.63, 0.85]).
Table~\ref{tab:weight} shows that variables $x_1$ to $x_4$ all had weights close to 0 as in the true setting none of them are distinguishable across clusters. While categorical variable $x_5$ had very low weight as it has very small differences across clusters in true setting.
Meanwhile, variables $x_6$ and $x_7$ which truly are associated with clustering had high weights (1.00 [1.00, 1.00]).

\begingroup
\renewcommand{\arraystretch}{1.4}
\begin{table}[H]
\footnotesize
\begin{adjustbox}{center}
\centering
\begin{tabular}{c c c c c}
\toprule
\textbf{Variable} & \multicolumn{4}{c}{\textbf{Weight, median (2.5th percentile, 97.5th percentile)}} \\ \midrule
 & Sim 1(a) & Sim 1(b) & Sim 2(a) & Sim 3 \\ \cmidrule(l){2-5}
\small{$x_1$} & 1.00 (1.00, 1.00) & 0.05 (0.04, 0.08) & 1.00 (0.95, 1.00) & 0.00 (0.00, 0.02) \\
\small{$x_2$} & 1.00 (1.00, 1.00) & 0.30 (0.11, 0.43) & 1.00 (1.00, 1.00) & 0.00 (0.00, 0.03) \\
\small{$x_3$} & 1.00 (1.00, 1.00) & 0.10 (0.05, 0.18) & 1.00 (1.00, 1.00) & 0.01 (0.00, 0.08) \\
\small{$x_4$} & 0.01 (0.01, 0.06) & 0.04 (0.03, 0.04) & 0.01 (0.01, 0.06) & 0.00 (0.00, 0.03) \\
\small{$x_5$} & 1.00 (1.00, 1.00) & 0.12 (0.06, 0.15) & 0.12 (0.08, 0.24) & 1.00 (1.00, 1.00) \\
\small{$x_6$} &  & 0.11 (0.06, 0.16) &  & 0.12 (0.09, 0.21) \\
\small{$x_7$} &  & 0.05 (0.04, 0.07) &  & 1.00 (1.00, 1.00) \\
\small{$x_8$} &  & 1.00 (1.00, 1.00) &  &  \\
\small{$x_9$} &  & 0.12 (0.08, 0.22) &  &  \\
\small{$x_{10}$} &  & 1.00 (1.00, 1.00) &  &  \\ \bottomrule
\end{tabular}
\caption{Obtained variable weights in different simulation settings}
\label{tab:weight}
\end{adjustbox}
\end{table}
\endgroup

In summary, the clustering performance of our Bayesian FMM is always the top performer across all simulation scenarios. In addition, our Bayesian FMM with variable selection is able to provide quantitative variable importance which is more informative than the dichotomous information we obtained using the HyDaP algorithm, especially for simulation 1(b).

\subsection{With existence of censored biomarker variables}
We evaluated the clustering performance of our proposed Bayesian FMM with variable selection under simulation setting 1(a), where existing methods performed well.
We compared the performances of proposed method with (1) our Bayesian FMM together with the naive method which uses half of lower detection limit to fill in undetected biomarker values; (2) the HyDaP algorithm together with the naive method; (3) PAM with Gower distance together with the naive method; and (4) K-prototypes together with the naive method.
We designed 4 scenarios based on the original simulation setting 1(a): (1) only variable $x_3$ and $x_4$ are censored and both with 20\% censoring; (2) only variable $x_3$ and $x_4$ are censored and both with 50\% censoring; (1) all continuous variables are censored and each with 20\% censoring; (2) all continuous variables are censored and each with 50\% censoring.
Clustering results are shown in Table~\ref{tab:censor} and obtained variable weights using proposed Bayesian FMM with variable selection are shown in Table~\ref{tab:censorw}.

\begingroup
\renewcommand{\arraystretch}{1.4}
\begin{table}
\footnotesize
\begin{adjustbox}{center}
\centering
\begin{tabular}{@{}ccccc@{}}
\toprule
\textbf{Clustering Method} & \multicolumn{4}{c}{\textbf{ARI, median (2.5th percentile, 97.5th percentile)}} \\ \midrule
 & \multicolumn{2}{c}{Censored variables: $x_3$, $x_4$} & \multicolumn{2}{c}{Censored variables: $x_1$ - $x_4$} \\
 & 20\%$^a$ & 50\%$^a$ & 20\%$^a$ & 50\%$^a$ \\ \cmidrule(l){2-5}
Bayesian FMM & 1.00 (0.98, 1.00) & 0.99 (0.96, 1.00) & 1.00 (0.97, 1.00) & 0.92 (0.42, 0.98) \\
Naive Bayesian FMM & 1.00 (0.98, 1.00) & 0.99 (0.44, 1.00) & 0.99 (0.65, 1.00) & 0.63 (0.46, 0.68) \\
HyDaP & 0.96 (0.92, 0.99) & 0.93 (0.89, 0.98) & 0.90 (0.84, 0.96) & 0.85 (0.75, 0.94) \\
PAM + Gower distance & 0.62 (0.53, 0.71) & 0.69 (0.59, 0.78) & 0.72 (0.57, 0.80) & 0.74 (0.65, 0.82) \\
K-prototypes & 0.99 (0.96, 1.00) & 0.98 (0.93, 1.00) & 0.90 (0.63, 0.97) & 0.60 (0.54, 0.66) \\ \midrule
\multicolumn{5}{l}{$^a$censoring proportion of each censored variable} \\
\bottomrule
\end{tabular}
\caption{Clustering performance with existence of censored biomarker variables}
\label{tab:censor}
\end{adjustbox}
\end{table}
\endgroup

In Table~\ref{tab:censor} we can observe that except the last column, our Bayesian FMM with variable selection performed the best.
When all the continuous variables have very high censoring proportions, our method performs typically well but with higher variability from simulation to simulation (ARI 0.92 [0.42, 0.98]).
Using Bayesian FMM with naive method to fill in censored values instead of using our embedded sampling approach, the performance is worse.
Meanwhile, performance of the HyDaP algorithm was consistently good, though not the best, across all scenarios.
When all the continuous variables have large censoring proportions, it can still yield high ARI with narrower percentile interval (ARI 0.85 [0.75, 0.94]).
PAM with Gower distance performed poorly in all scenarios.
Performance of K-prototypes is satisfactory when only a few continuous variables are censored, but becomes worse when all continuous variables are censored.
In terms of variable weights, Table~\ref{tab:censorw} indicates that our Bayesian FMM with embedded sampling approach yields weights that reflect the true setting for all scenarios.

\begingroup
\renewcommand{\arraystretch}{1.4}
\begin{table}[H]
\footnotesize
\begin{adjustbox}{center}
\centering
\begin{tabular}{@{}ccccc@{}}
\toprule
\textbf{Weight} & \multicolumn{4}{c}{\textbf{ARI, median (2.5th percentile, 97.5th percentile)}} \\ \midrule
 & \multicolumn{2}{c}{Censored variables: $x_3$, $x_4$} & \multicolumn{2}{c}{Censored variables: $x_1$ - $x_4$} \\
 & 20\%$^a$ & 50\%$^a$ & 20\%$^a$ & 50\%$^a$ \\ \cmidrule(l){2-5}
$x_1$ & 1.00 (1.00, 1.00) & 1.00 (1.00, 1.00) & 1.00 (1.00, 1.00) & 1.00 (1.00, 1.00) \\
$x_2$ & 1.00 (1.00, 1.00) & 1.00 (1.00, 1.00) & 1.00 (1.00, 1.00) & 1.00 (1.00, 1.00) \\
$x_3$ & 1.00 (1.00, 1.00) & 1.00 (1.00, 1.00) & 1.00 (1.00, 1.00) & 1.00 (1.00, 1.00) \\
$x_4$ & 0.03 (0.02, 0.14) & 0.01 (0.01, 0.05) & 0.02 (0.01, 0.10) & 0.01 (0.01, 0.04) \\
$x_5$ & 1.00 (1.00, 1.00) & 1.00 (1.00, 1.00) & 1.00 (1.00, 1.00) & 1.00 (1.00, 1.00) \\ \midrule
\multicolumn{5}{l}{$^a$censoring proportion of each censored variable} \\
\bottomrule
\end{tabular}
\caption{Obtained variable weights with existence of censored biomarker variables}
\label{tab:censorw}
\end{adjustbox}
\end{table}
\endgroup

In summary, when most continuous variables are censored with high probability, we suggest using the HyDaP algorithm since it would provide robust performance under this scenario. In all other scenarios, our proposed Bayesian FMM with variable selection is preferred as it would yield the best results.

\section{Application}
\label{sec:real2}

We analyzed the SENECA data and assumed that the data contains three latent clusters, as shown in ~\citep{2019arXiv190502257W}. Variables used in SENECA are listed in the appendix.

Due to the assumption of conditional independence, namely variables are independent with each other conditional on cluster labels, we dropped variables heart rate, blood sodium level (Na), aspartate aminotransferase (AST), and blood urea nitrogen (BUN) because of their high correlations to other variables in the dataset. The goals of our analysis are to obtain quantitative weights for all variables in the analytic dataset and to assess whether the identified clusters are related to those identified by the HyDaP algorithm.

Our hyper-parameters used in the analysis are as follows: $\bm{\delta} =(0.33,0.33,0.33)^T$ was used assuming that we do not have much information on $\bm{\tau}$. We set $\mu_A=0$ and $\sigma_A^2=1,000$ assuming that we do not have much information on $A_{1m}$. To set $a_{\Delta_0}=2$ and $b_{\Delta_0}=0.005$ as we believe that $\sigma_{\Delta_0}^2$ should be a small number so that the corresponding prior for $\mu_{mg}$ is a \textit{spike} one. We set $\sigma_{\Delta_1}^2=100$ as we believe that $\sigma_{\Delta_1}^2$ should be a large number so that the corresponding prior for $\mu_{mg}$ is a \textit{slab} one. We also let $\tilde{a}=2$ and $\tilde{b}=1$ assuming that we do not have much information on $\gamma_m$. $\bm{\alpha}_{\Delta_1}=1$ so that the corresponding prior for $\theta_{mg}$ is a \textit{slab} one. We set $\bm{\alpha}_{\Delta_0}=(10.2, 10)^T $ for gender and $\bm{\alpha}_{\Delta_0}=(15.5, 2.4,2.1)^T $ for race based on their marginal distributions. We would like some shrinkage on both continuous and categorical variables, so we set $a_{p_1}=1$, $b_{p_1}=2$, $a_{p_2}=1$, and $b_{p_2}=2$.

\begingroup
\renewcommand{\arraystretch}{1.4}
\begin{table}[H]
\footnotesize
\begin{adjustbox}{center}
\centering
\begin{tabular}{@{}lc|cccc@{}}
\toprule
\multicolumn{2}{l|}{\multirow{2}{*}{}} & \multicolumn{4}{c}{\textbf{HyDaP}} \\
\multicolumn{2}{l|}{} & Cluster 1 & Cluster 2 & Cluster 3 & Cluster 4 \\ \midrule
\multirow{3}{*}{\textbf{Bayesian FMM}} & Cluster 1 & 5083 (50.9\%) & 2614 (26.2\%) & 1729 (17.3\%) & 557 (5.6\%) \\
& Cluster 2 & 1241 (16.7\%) & 2411 (32.4\%) & 3073 (41.3\%) & 722 (9.7\%) \\
& Cluster 3 & 156 (5.7\%) & 238 (8.6\%) & 730 (26.5\%) & 1635 (59.3\%) \\ \bottomrule
\end{tabular}
\caption{A comparison between clusters identified by the proposed Bayesian FMM and the HyDaP algorithm}
\label{tab:cross}
\end{adjustbox}
\end{table}
\endgroup

By applying our proposed Bayesian FMM, we obtained 3 clusters with sample sizes $9,983$, $7,447$, and $2,759$, respectively. Table~\ref{tab:cross} shows these results cross tabulating with the 4 clusters that were identified using the HyDaP algorithm. Cluster 1 and Cluster 3 identified by the Bayesian FMM are similar to Cluster 1 and Cluster 4 identified by the HyDaP algorithm, respectively. Cluster 2 of the Bayesian FMM is similar to Clusters 2 and 3 altogether from the HyDaP algorithm. The results show that the 3 clusters identified by using the proposed Bayesian FMM are consistent with those identified from the HyDaP algorithm.

Table~\ref{tab:senecaw} summarizes the clustering weight of each variable obtained from the Bayesian FMM. None of the variables had driving influences on the final clustering results as their weights were low. This finding is consistent with what we found using the HyDaP algorithm. An advantage of using our proposed Bayesian FMM in ranking variable importance is its ability to provide a real-valued weight for each variable, not just dichotomizing certain variables as important or not. In SENECA, we found that ESR, troponin, lactate, sex, albumin, bicarbonate, GCS, and INR had relatively high weights, which mean that these variables contributed more in forming the 3 clusters than other variables did. Such information provides more details than the general importance obtained from the HyDaP algorithm. This is especially useful for clinical data which usually do not contain any dominant variables in terms of clustering.

\begingroup
\renewcommand{\arraystretch}{1.4}
\begin{table}
\footnotesize
\begin{adjustbox}{center}
\centering
\begin{tabular}{@{}cc|cc|cc@{}}
\toprule
\textbf{Variable} & \textbf{Weight} & \textbf{Variable} & \textbf{Weight} & \textbf{Variable} & \textbf{Weight} \\ \midrule
Age & 0.05 & GCS & 0.12 & CRP & 0.10 \\
Temperature & 0.05 & Elixhauser score & 0.04 & INR & 0.12 \\
Systolic blood pressure & 0.11 & White blood cell & 0.06 & Glucose & 0.05 \\
Respiration rate & 0.08 & Bands & 0.08 & Platelets & 0.05 \\
Albumin & 0.13 & Creatinine & 0.05 & SaO2 & 0.05 \\
Cl & 0.07 & Bilirubin & 0.09 & PaO2 & 0.08 \\
ESR & 0.30 & Troponin & 0.27 & Gender & 0.20 \\
Hemoglobin & 0.06 & Lactate & 0.22 & Race & 0.06 \\
Bicarbonate & 0.12 & ALT & 0.11 & & \\ \midrule
\multicolumn{6}{l}{Abbreviation: ESR: Erythrocyte sedimentation rate; GCS: Glasgow coma scale;} \\
\multicolumn{6}{l}{ALT: Alanine aminotransferase; CRP: C-reactive protein; INR: International normalized ratio;} \\
\multicolumn{6}{l}{SaO2: Oxygen saturation; PaO2: Partial pressure of oxygen} \\
\bottomrule
\end{tabular}
\caption{Clustering weights of the SENECA variables using proposed Bayesian FMM}
\label{tab:senecaw}
\end{adjustbox}
\end{table}
\endgroup

\begin{figure}
\centering
\includegraphics[width=0.9\textwidth]{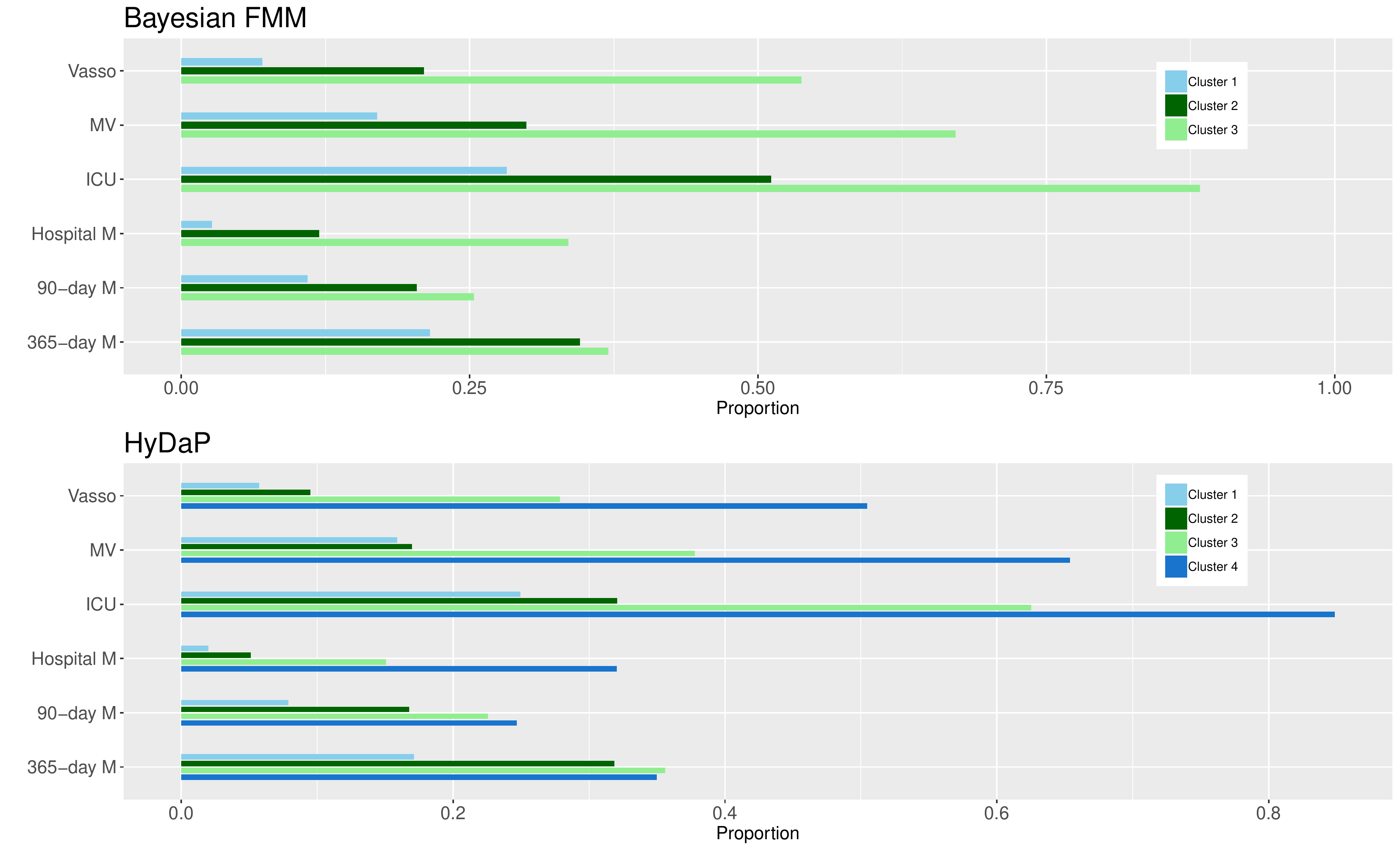}
\caption{Distributions of selected clinical endpoints across clusters identified by the proposed Bayesian FMM and the HyDaP algorithm}
\label{wsenecak3}
\end{figure}

Table~\ref{tab:wclinical} and Figure~\ref{wsenecak3} show the distributions of clinical endpoints across the 3 clusters obtained by Bayesian FMM with variable selection.
We can observe that Cluster 1 has the lowest proportion for all clinical endpoints while Cluster 2 has the second lowest ones. Cluster 3 has the highest proportions. We observe that these also correspond to the results obtained in the HyDap algorithm but our method is able to provide quantitative weights of all involved variables so that we have better knowledge about their importance and better interpretation of clustering results.

\begingroup
\renewcommand{\arraystretch}{1.4}
\begin{table}
\footnotesize
\begin{adjustbox}{center}
\centering
\begin{tabular}{@{}llllll@{}}
\toprule
\multicolumn{1}{c}{Clinical Endpoints} & \multicolumn{1}{c}{All Clusters} & \multicolumn{1}{c}{Cluster 1} & \multicolumn{1}{c}{Cluster 2} & \multicolumn{1}{c}{Cluster 3} & \multicolumn{1}{c}{\textit{p}-value} \\ \midrule
& \multicolumn{5}{c}{Bayesian FMM with variable selection} \\ \cmidrule(l){2-6}
Cluster size & 20189 & 9984 (49.4\%) & 7447 (36.9\%) & 2759 (13.7\%) & \\
Admitted to ICU & 9063 (44.9\%) & 2817 (28.2\%) & 3809 (51.1\%) & 2437 (88.3\%) & \textless{}0.001 \\
Mechanical Ventilation & 5773 (28.6\%) & 1694 (17.0\%) & 2227 (29.9\%) & 1852 (67.1\%) & \textless{}0.001 \\
Vasopressor & 3755 (18.6\%) & 703 (7.0\%) & 1568 (21.1\%) & 1484 (53.8\%) & \textless{}0.001 \\
In-hospital mortality & 2081 (10.3\%) & 267 (2.7\%) & 889 (11.9\%) & 926 (33.6\%) & \textless{}0.001 \\
\begin{tabular}[c]{@{}l@{}}90-day mortality$^a$ \\ (exclude in-hospital mortality)\end{tabular} & 2758 (14.2\%) & 1029 (11.0\%) & 1286 (20.4\%) & 443 (25.4\%) & \textless{}0.001 \\
\begin{tabular}[c]{@{}l@{}}365-day mortality$^b$\\ (exclude in-hospital mortality)\end{tabular} & 5043 (27.9\%) & 2096 (21.6\%) & 2268 (34.6\%) & 679 (37.0\%) & \textless{}0.001 \\ \midrule
\multicolumn{6}{l}{$^a$Total number is 17,432 after excluding in-hospital death and missing.} \\
\multicolumn{6}{l}{$^b$Total number is 18,107 after excluding in-hospital death.} \\ \bottomrule
\end{tabular}
\caption{Distributions of selected clinical endpoints across clusters identified by the Bayesian FMM}
\label{tab:wclinical}
\end{adjustbox}
\end{table}
\endgroup

\section{Discussion}
\label{sec:dis2}
Clustering has received a lot of attention and been applied in various areas these days. However, clustering methods that can handle mixed types of variables (both continuous and categorical) are still limited. Finite mixture model is a branch of clustering methods that is able cluster mixed data. But most existing FMM frameworks with variable selection are limited to single data type. Therefore, the goal of this paper is to develop FMM that can cluster data with mixed types of variables and perform variable selection in clustering.

We proposed a Bayesian FMM that can simultaneously cluster variables with mixed types, calculate variable weights, as well as handle censored biomarker variables. In this method we apply a Bayesian framework in order to bypass the limitations in the EM algorithm which is the standard estimation method for a FMM. In addition to identifying clusters, our model can provide real-valued variable weights which are more informative than a dichotomy of a variable being important vs. not in clustering. In addition, our proposed method is able to handle censored biomarker variables through recovering their underlying distributions. For a dataset of variables with mixed types but without censoring, our proposed Bayesian FMM performs better than the HyDaP algorithm and other existing methods across various simulation settings. If censored variables exist in the data, proposed Bayesian FMM with embedded sampling approach performs better than other clustering algorithms with naive fill-in methods. However, when variables have high censoring proportions, the proposed Bayesian FMM with embedded sampling approach may not consistently outperform other approaches. In this case, the HyDaP algorithm with ad-hoc imputations provides better and robust results. Under all scenarios, our proposed Bayesian FMM with variable selection is able to provide reasonable weights of all variables.

Users of the proposed Bayesian FMM model need to be aware of certain limitations. Same as finite mixture models, the proposed model also has the disadvantage of unverifiable distributional assumptions. Besides, the model assumes conditional independence (i.e., variables are independent conditional on cluster membership) so it may not perform well when variables are subject to within-cluster correlations. In computations, because that we need to specify the values for all hyper-parameters before running the algorithm, the computation time for the proposed model is usually longer than that for the EM algorithm.

Future work can be done on the top of the development of this paper. For the Bayesian FMM with variable selection, the framework can be extended to account for more distributions. Approaches of handling censored variables can also be further developed.


\bibliographystyle{apalike} \bibliography{ref}

\clearpage

\appendix

\section{\\Variables used in SENECA data analysis}

Age\\
Gender: categorical variable; 2 levels (male/female)\\
Race: categorical variables; 3 levels (white/black/hispanic)\\
Maximum temperature within 6 hours of ER presentation\\
Maximum heart rate within 6 hours of ER presentation\\
Minimum systolic blood pressure within 6 hours of ER presentation\\
Maximum respiration rate within 6 hours of ER presentation\\
Maximum albumin within 6 hours of ER presentation\\
Maximum Cl within 6 hours of ER presentation\\
Maximum erythrocyte sedimentation rate (ESR) within 6 hours of ER presentation\\
Maximum hemoglobin within 6 hours of ER presentation\\
Maximum bicarbonate within 6 hours of ER presentation\\
Maximum Sodium within 6 hours of ER presentation\\
Minimum Glasgow Coma Scale (GCS) within 6 hours of ER presentation\\
Elixhauser Score\\
Maximum white blood cell within 6 hours of ER presentation\\
Maximum bands within 6 hours of ER presentation\\
Maximum creatinine within 6 hours of ER presentation\\
Maximum bilirubin within 6 hours of ER presentation\\
Maximum troponin within 6 hours of ER presentation\\
Maximum lactate within 6 hours of ER presentation\\
Maximum alanine aminotransferase (ALT) within 6 hours of ER presentation\\
Maximum aspartate aminotransferase (AST) within 6 hours of ER presentation\\
Maximum C-reactive protein within 6 hours of ER presentation\\
Maximum international normalized ratio (INR) within 6 hours of ER presentation\\
Maximum glucose within 6 hours of ER presentation\\
Maximum Platelets within 6 hours of ER presentation\\
Maximum blood urea nitrogen (BUN) within 6 hours of ER presentation\\
Oxygen saturation (SaO2)\\
Minimum partial pressure of oxygen (PaO2) within 6 hours of ER presentation

\end{document}